\pdfoutput=1

\documentclass[11pt]{article}
\usepackage{subcaption}
\usepackage{tikz} 
\usepackage{tabularx}
\usepackage{graphicx}
\usepackage{booktabs}
\usepackage{multirow}
\usepackage{pgfplots}
\usepackage{algorithm}
\usepackage{algorithmicx}
\usepackage{algpseudocode}
\usepackage{xcolor}

\usepackage[preprint]{acl}

\newcommand{\TargetVenue}{EMNLP}

\newcommand{\FrameworkName}{TIDE}
\newcommand{\FrameworkNameShort}{TIDE}
\newcommand{\FrameworkExplaned}{criteria optimizing with \textbf{T}r\textbf{I}al and \textbf{DE}bate}

\newcommand{\MetaSolution}{\textit{Guider}}
\newcommand{\Solution}{\textit{Solver}}
\newcommand{\RetryTimesMarkPure}{n_{trial}}
\newcommand{\RetryTimesMark}{$\RetryTimesMarkPure$}
\newcommand{\IterationMarkPure}{n_{Iter}}
\newcommand{\IterationMark}{$\IterationMarkPure$}
\newcommand{\DatasetName}{CEAMC}

\newcommand{\RetryMark}{Trial}
\newcommand{\DebateMark}{Debate}

\usepackage{times}
\usepackage{latexsym}

\usepackage[T1]{fontenc}

\usepackage[utf8]{inputenc}

\usepackage{microtype}

\usepackage{inconsolata}

\usepackage{graphicx}

\title{Towards Robust Argumentative Essay Understanding via TIDE: An Interactive Framework with \RetryMark~and \DebateMark}

\author{
    Zheqin Yin\textsuperscript{$1$},
    Yupei Ren\textsuperscript{$1$,$2$,$3$},
    Yadong Zhang\textsuperscript{$1$},
    Yujiang Lu\textsuperscript{$1$},
    \textbf{Man Lan\textsuperscript{$1$,$2$,$3$}\thanks{\ \ Corresponding author.}}
    \\
    \textsuperscript{1}School of Computer Science and Technology, East China Normal University \\
    \textsuperscript{2}Shanghai Institute of Artificial Intelligence for Education, East China Normal University\\
    \textsuperscript{3}Lab of Artificial Intelligence for Education, East China Normal University \\
    \texttt{zqyin@stu.ecnu.edu.cn, mlan@cs.ecnu.edu.cn} \\
}
\begin{document}
\maketitle
\begin{abstract}
Argumentative essays serve as a vital medium for assessing critical thinking and reasoning skills, yet there is limited works on accurately understanding and evaluating such texts via prompt. In this work, we propose \textbf{\FrameworkName}, a novel framework designed to improve criteria-based prompt optimization for argument-related tasks by integrating \textbf{T}r\textbf{I}al and \textbf{DE}bate mechanism. Our method addresses key limitations of criteria-based prompt optimizing by mitigating the influence of noisy training data and enhancing optimization stability. We evaluate \FrameworkName~on three core tasks: Automated Essay Scoring, Argument Component Detection, and Argument Relation Identification. Results demonstrate that our framework improves performance across tasks. These findings underscore the potential of combining prompt-based methods for advanced argument understanding.
\end{abstract}

\section{Introduction}

Argumentative essays, as a genre of academic writing, serve as tangible artifacts that reflect ones' abilities to construct, articulate, and defend coherent arguments \cite{Drury2019ArgumentPF,Ulfa2023ArgumentativeEP}. Understanding and evaluating argumentative essays, i.e. conducting argument mining, not only provides a window into the study of argumentative thinking but also offers a practical pathway for promoting the development of this crucial cognitive skill \cite{Lu2021InfusingCT,Mombaers2024LearningFC}.

In recent years, the predominant approaches in argument mining have focused on training pretrained language models or fine-tuning large language models (LLMs), both of which have demonstrated strong performance on this task \cite{Favero2025LeveragingSL,Wang2020ArgumentationMO}. However, current research in this field has seen limited exploration of novel frameworks built upon prompt-based methods, despite their advantages in terms of simplicity and flexibility. With the emergence of advanced reasoning models such as DeepSeek-R1 \cite{DeepSeekAI2025DeepSeekR1IR}, prompt-based approaches are becoming an increasingly promising alternative, warranting deeper investigation within the context of argument mining.

\begin{figure*}[t]
    \centering
    \begin{subfigure}[b]{0.43\textwidth}
        \includegraphics[width=\textwidth]{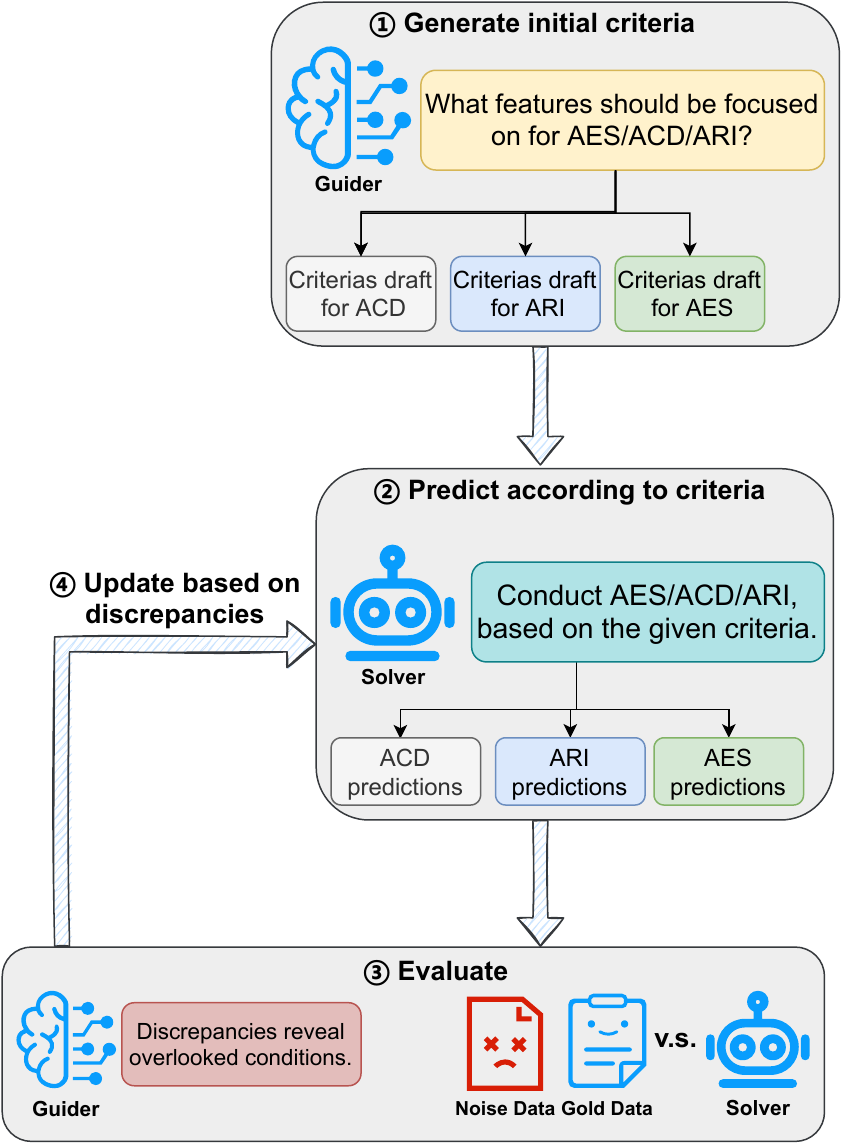}
        \caption{The process of criteria-based prompt optimizing.}
        \label{fig:basic_ver}
    \end{subfigure}
    \begin{subfigure}[b]{0.55\textwidth}
        \includegraphics[width=\textwidth]{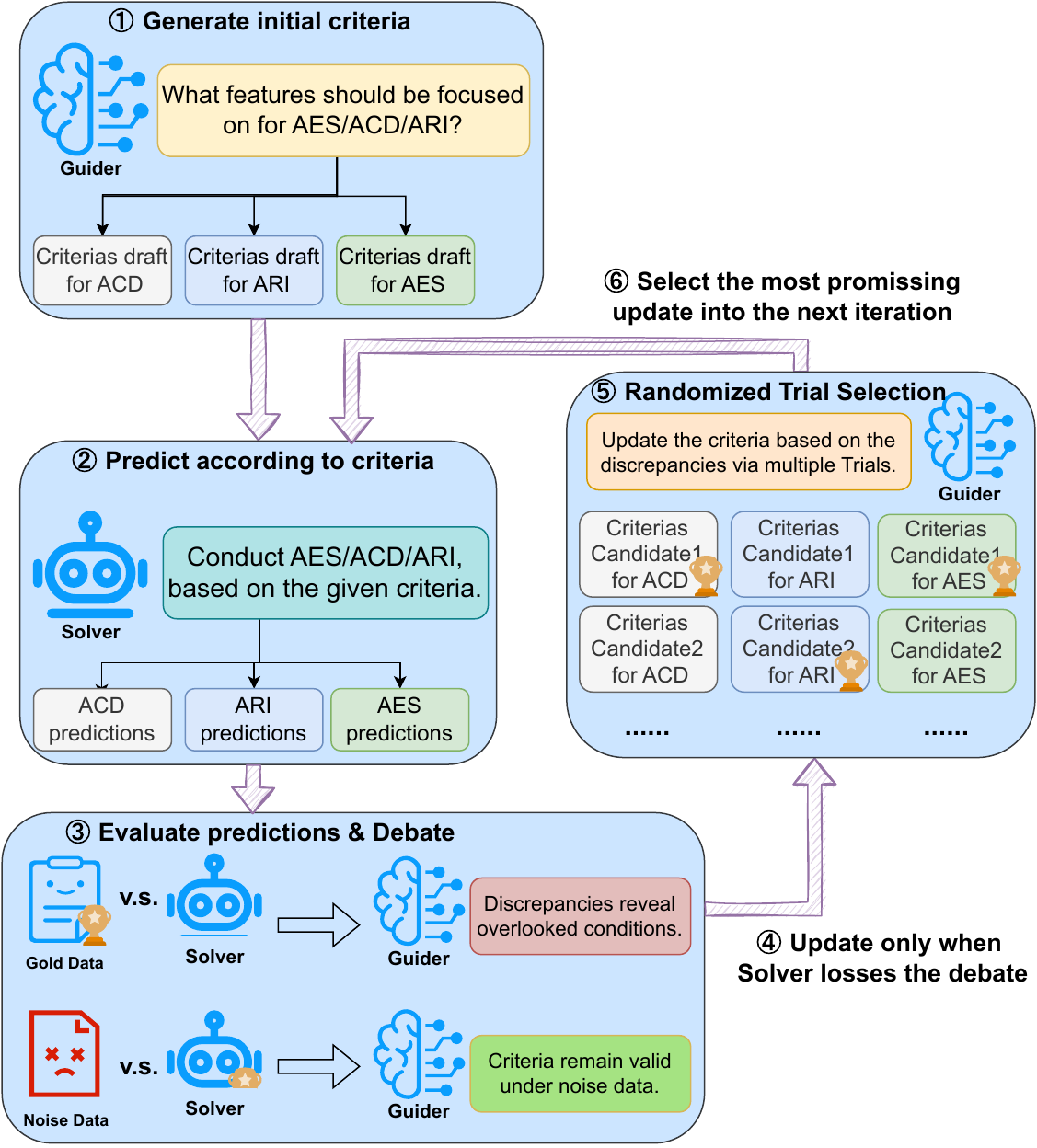}
        \caption{Our proposed framework \FrameworkName}
        \label{fig:overview}
    \end{subfigure}
    \caption{The overview of criteria-based prompt optimizing (Figure \ref{fig:basic_ver}) and our proposed framework \FrameworkName~(Figure \ref{fig:overview}), where \DebateMark~and Randomized Trial Selection is employed to enhance the optimizing process.}
    \label{fig:overview_all}
\end{figure*}

Recent studies demonstrate that criterion-based prompt optimization methods, as illustrated in Figure \ref{fig:basic_ver}, can achieve optimization through implicit task signals in training data without directly modifying the prompt text of LLMs \cite{yang2023large,Liu2023CalibratingLE}. While this approach aims to emulate human-like abstraction of inferential rules \cite{barwise1993everyday} from observations, existing methods often suffer from critical limitations. Specifically, this approach refines the initial criteria in a gradient-free, iterative manner, which lacks performance guarantees and may be overly sensitive to noisy or unrepresentative samples. To address these challenges, we propose \FrameworkExplaned~(\FrameworkNameShort), a novel framework that leverages Randomized Trial Selection and \DebateMark~to enhance the performance of understanding argumentative essays. The overview of \FrameworkNameShort~can be found in Figure \ref{fig:overview}. First, to mitigate the adverse influence of noisy or unrepresentative training data, we introduce a \DebateMark~process that allows the current criteria to "defend" itself. Furthermore, to enhance the quality of each refinement step, we introduce a Randomized \RetryMark~Selection mechanism, which explores multiple candidate updates and selects the most promising one, thereby improving both stability and convergence of the optimization process.

We evaluate our \FrameworkNameShort~across three representative tasks—Automated Essay Scoring (\textbf{AES}), Argument Component Detection (\textbf{ACD}), and Argument Relation Identification (\textbf{ARI}), where the results demonstrate the effectiveness of our framework. We find that task complexity should guide the configuration of these components, with simpler tasks like AES benefiting from minimal refinement but need more debate, while more complex tasks like ARI requiring deeper iterative processing for optimal outcomes. Furthermore, we highlight that the reasoning ability of the base models is essential for fully realizing the potential of \FrameworkName~for DeepSeek series models by comparing DeepSeek-R1 and DeepSeek-V3 \cite{DeepSeekAI2024DeepSeekV3TR}. Substituting these models with weaker alternatives results in substantial performance degradation across all tasks. This work aims to provide a comprehensive understanding of the framework's capabilities and its implications for improving argumentative essay understanding and evaluating tasks. In summary, our main contributions are as follows:
\begin{itemize}
    \item We conduct a systematic study on o1-like models, specially DeepSeek-R1, in understanding human argumentative thinking, grounded in three representative tasks: AES, ACD, and ARI.
    
    \item We propose a novel prompt optimization framework, \FrameworkName, which integrates Randomized Trial Selection and \DebateMark~to significantly enhance model performance in argument understanding and evaluation across the three tasks.
    
    \item We empirically analyze the role of reasoning within \FrameworkName~and demonstrate that reasoning ability is essential to fully unlocking the argumentative potential of DeepSeek-series.
\end{itemize}

\section{Related Works}

\subsection{Argument Mining}
In the domain of argumentation-related tasks, prior research has predominantly focused on pretrained language models such as BERT \cite{Devlin2019BERTPO} as its variants \cite{Sazid2022AUR,Cheng2020ArgumentPE}. In recent times, there has been an emerging trend to delve into the performance of LLMs in pertinent tasks \cite{Favero2025LeveragingSL,Gorur2024CanLL}. For instance, \citet{Favero2025LeveragingSL} investigated the application of open-source models such as Qwen and LLaMA for argument segmentation and argument type classification within educational settings. On the other hand, \citet{Gorur2024CanLL} explored the performance of LLaMA and Mistral models with varying parameter scales on the task of relation prediction. 

To the best of our knowledge, most existing studies have concentrated on either fine-tuning pretrained language models or developing task-specific adaptations of LLMs, leaving a gap in developing new forms of prompting strategies for argumentation-related tasks.

\subsection{Debate by LLMs}

There has been growing interest in exploring the interaction between LLMs and the concept of debate \cite{Liang2023EncouragingDT,Irving2018AISV}. Some researchers treat debate as a scenario to probe and evaluate relevant capabilities of LLMs. For example, \citet{He2024DecomposingAE} employed LLMs to compose argumentative essays. In addition, \citet{Arnesen2024TrainingLM} assigned them roles as participants and trained them to win debates, while \citet{Liang2024DebatrixMD} positioned them as judges to assess the quality of arguments from three dimensions. 

On the other hand, there are researchers incorporating debate as a functional module within broader system architectures. They treat LLMs as argumentative collaborators more than just answer providers \cite{musi2025toward}. For instance, \citet{Wu2024LLMBasedER} employed debate between different psychologist-agents to enhance empathetic response of their system in the domain of psychological diagnostics. Similarly, in the context of chain-of-thought (CoT) prompting, debate mechanisms have been introduced to facilitate more robust math capability \cite{wan2024cot}.

\subsection{Prompt Optimizing Strategies}

Rather than relying on manually crafted prompts, which can be time-consuming, suboptimal, and difficult to generalize across tasks, an alternative approach is to leverage the LLM itself to generate or refine prompts. For example, APE \cite{Zhou2022LargeLM} uses LLMs to propose candidate prompts, which are then evaluated and selected based on their performance on a given task. This approach achieves human-level performance on various tasks with minimal human input. Similarly, \citet{Yang2023LargeLM} introduces Optimization by PROmpting (OPRO), a method leveraging LLMs as optimizers. OPRO describes optimization tasks in natural language and iteratively generates new solutions based on previously evaluated ones. This approach is demonstrated in prompt optimization, where prompts are optimized to maximize task accuracy. Different from optimize the prompt itself, \citet{Liu2023CalibratingLE} employed the in-context learning capability of LLMs to capture criteria from annotated examples, which is then assessed and refined through self-correction, ultimately yielding a calibrated version of criteria. 
\section{Method}

\subsection{Criteria-based prompt optimizing}
\label{sec:basic_ver}
Criteria are central to prompts for guiding LLMs toward better performance. For instance, users may include directives such as "…carefully analyze coherence, structure, and argumentation before giving the final score…" to improve alignment in scoring argumentative essays. However, how should the LLM understand coherence, and, how much weight should be placed on it? Given the typical absence of detailed rubric in user-provided prompts or annotation guidelines, criteria-based prompt optimization is used to generate and refine well-specified criteria, thereby enhancing the guidance for LLM predictions.

An overview of criteria-based prompt optimizing is illustrated in Figure \ref{fig:basic_ver}. Initially, \MetaSolution~produces an initial version of criteria, denoted as $c_0$, which in our context serves as a guideline for understanding and assessing argumentative essays. Then \Solution~is employed to perform the corresponding task with $c_0$, producing a predicted output $\hat{y}$ given an input $x\in\mathcal{X}$. By comparing the predicted output $\hat{y}$ with the ground truth $y$, \MetaSolution~is guided to update the criteria from $c_0$ to $c_1$. This updated critiera is then evaluated on the training data again, and the refinement process continues. 

However, several limitations exist in this vanilla process. First, since no gradients are updated throughout this pipeline, there is no guarantee that the updated criteria $c_{i+1}$ will lead to improved performance of \Solution~compared to the previous version $c_i$. Moreover, forcing \MetaSolution~to update the criteria solely based on observed discrepancies may be problematic, as it could be influenced by noisy or unrepresentative data, potentially degrading overall performance.

\subsection{Overall Framework of \FrameworkNameShort}

We propose \FrameworkExplaned~(\FrameworkNameShort), which leverages \DebateMark~and Randomized Trial Selection to improve criteria-based prompt optimization as shown in Figure \ref{fig:overview}. \FrameworkNameShort~ begins by initializing the criteria draft $c_0$ by \MetaSolution~and iterates over batches $B$ of the training dataset \(D_{train}\). For each batch, \Solution~generates predictions, computes discrepancy (see Section \ref{sec:mct}), and identifies the sample with the largest discrepancy. A debate is then conducted between the predicted value $\hat{y}$ and the ground truth $y$ to determine whether an update is necessary (see Section \ref{sec:debate}). If the $\hat{y}$ wins the debate, the process proceeds into the next iteration without update. Otherwise, the algorithm generates multiple candidate updates and selects the one with the minimal error on the same batch $B$ as the final update. Algorithm \ref{alg:tide} presents detailed \FrameworkNameShort~algorithm. 

\subsection{Debate module}
\label{sec:debate}

\begin{algorithm}[H]
\caption{\FrameworkNameShort}
\begin{algorithmic}[1]
\Require  
Training dataset $D_{train}$, max iteration \IterationMark, batch size $bsz$, number of \RetryMark~\RetryTimesMark

\Ensure Refined criteria $c$

\State Generate initial criteria $c_{current} \gets c_0$

\For{$i = 0$ to \IterationMark}
    \For{Batch $B \in D_{train}$}
        \State Employing $c_{current}$ to generate predictions $\hat{y} = \hat{y}_1, \cdots, \hat{y}_{bsz}$
        \State Compute discrepancy for batch $b = b_1, \cdots, b_{bsz}$
        \State Pick the sample with the largest discrepancy $i_{max}, b_{i_{max}} = \max(b)$
        \State Debate between $\hat{y}_{i_{max}}$ and $y$    
        
        \If{$y$ wins} 
            \State Generate \RetryTimesMark~candidate updates $\hat{c}_{i+1} = \hat{c}_{i+1}^1, \cdots, \hat{c}^{\RetryTimesMarkPure}_{i+1}$ with $B[i_{max}]$
            \State The minimal discrepancy $b_{min} \gets \infty$
            \State The final update $c_{final} \gets \textbf{None}$
            \For{$\hat{c} \in \hat{c}_{i+1}$}
                \State Compute discrepancy of $B$ using $\hat{c}$ to get $b^{\hat{c}}$
                \If{$max(b^{\hat{c}}) \leq b_{min}$}
                    \State $b_{min} \gets max(b^{\hat{c}})$
                    \State $c_{final} \gets \hat{c}$
                \EndIf
            \EndFor
            \State $c_{current} \gets c_{final}$
        \EndIf
    \EndFor
\EndFor
\State \Return $c_{current}$
\end{algorithmic}
\label{alg:tide}
\end{algorithm}

Many studies have demonstrated that debate serves as an effective mechanism for enhancing the truthfulness of system-generated responses, which is primarily because LLMs have been shown to struggle when attempting to defend false or inaccurate claims \cite{Michael2023DebateHS, Khan2024DebatingWM,Du2023ImprovingFA}. Leveraging this characteristic, and recognizing the natural compatibility between debate structures and argumentative tasks, we introduce a debate-based process aimed at mitigating the impact of noisy data in argumentative contexts. We further extend the evaluation process by incorporating a simple internal debate mechanism, which compares the predicted output with the gold reference. This simulated debate in turn serves to control the iteration condition, allowing the model to refine its outputs more selectively and robustly. 

Concretely, when \Solution~produces an incorrect prediction $\hat{y}$ based on a given criterion $c_i$, a debate is initiated between the predicted output $\hat{y}$ and the correct label $y$. In this setting, the "speech" for $\hat{y}$ is the explanation originally generated by \Solution~at prediction time, while the "speech" for $y$ is constructed by prompting the LLM to generate a plausible explanation supporting the correct label. A LLM-based judge is then employed to assess which explanation better aligns with the annotation criterion (differ according to specific dataset). The update will only initialize when $\hat{y}$ won the debate, or the process would proceed to next iteration. This debate mechanism allows \MetaSolution~to reduce the influence of potentially noisy annotations present in the training data, as their explanation would be less convincing, thereby enhancing the overall robustness of the framework, resulting in the training dynamic illustrated in Figure \ref{fig:discrepancy_comparison}.

\subsection{Randomized Trial Selection}
\label{sec:mct}
We introduced randomized trial selection upon updating criteria. Specifically, given the previous criteria $c_{i}$, multiple candidate updates, denoted as $\hat{c}_{i+1}^0, \hat{c}_{i+1}^1, \cdots$, are generated, where each sample serves as an estimate of the potentially optimal update from $c_{i}$. More formally, an update is generated via $\hat{c}_{i+1}=\pi_\theta(\mathcal{T}, c_i)$, where $\mathcal{T}$ is the prompt template, $\pi_\theta$ is \MetaSolution~with parameter $\theta$. We repeat this generation for \RetryTimesMark~times to get different candidates. 

Among the generated candidates, \textit{discrepancy} between the prediction and the latent criteria embedded in the data is computed to identify the most promising update. We measure the discrepancy via the error predictions, which may differ according to each task. Specifically, for ACD, we regard the number of predicted labels that do not match the ground truth labels, while the absolute difference between predicted and labeled scores for AES. For ARI, which is rather complicated, we consider both the error rate of the identification of related pairs and the prediction of the relationship categories between them. Additionally, we introduce a penalty mechanism that imposes a higher penalty when the model fails to recognize the existence of any relationship between a pair of sequences, prior to predicting the exact type of relationship. Details of how the discrepancy is computed for each task can be found in Appendix \ref{adx:discrepancy}. The candidate with the lowest score is then selected as the final update for $c_{i+1}$.

\section{Experiment Setup}

\begin{table*}[htbp]
    \centering
    \begin{tabular}{l|cc|cc|cc|cc}
        \toprule
        \multirow{3}{*}{Method} & \multicolumn{4}{c|}{\DatasetName} & \multicolumn{4}{c}{AEE} \\
        \cline{2-9}
        & \multicolumn{2}{c|}{ACD}& \multicolumn{2}{c|}{ARI} & \multicolumn{2}{c|}{ACD}& \multicolumn{2}{c}{ARI}\\
        & Micro & Macro & Micro & Macro & Micro & Macro & Micro & Macro\\
        \midrule
        ICL & 62.75 & 49.21 & 6.50 & 8.47 & 84.06 & 82.00 & 51.42 & 37.19 \\
        CoT & 62.27 & 48.83 & 8.23 & 9.64 & 85.07 & 82.78 & 56.05 & 38.76\\
        Calibrate & 63.37 & 50.29 & 10.76 & 9.97 & 82.25 & 59.95 & 58.06 & 40.99 \\
        \hline
        Criteria-based & 67.28 & 48.69 & 9.27 & 10.11 & 82.70 & 81.64 & 60.13 & 40.18 \\
        \FrameworkNameShort & \textbf{69.57} & \textbf{56.85} & \textbf{16.05} & \textbf{15.39} & \textbf{85.35} & \textbf{83.39} & \textbf{62.83} & \textbf{43.83} \\
        \bottomrule
    \end{tabular}
    \caption{Performance comparison with different baselines on ACD and ARI, both from \DatasetName~dataset. All results in this table are presented in percent format(\%).}
    \label{tab:main_results}
\end{table*}

\subsection{Task Format}

In this work, we mainly focus on three representative argument-related tasks:

\textbf{Automated Essay Scoring (AES)}: This task involves assigning an overall score to an input argumentative essay ranged from 1 to 5, based on the quality, coherence, and other relevant aspects of the argumentation. 

\textbf{Argument Component Detection (ACD)}: Given an argumentative essay $D$ consisting of $n$ discourse units, i.e., $ D = [s^1, ..., s^n]$, the task requires predicting the fine-grained type of each sentence, resulting in a label sequence $\mathcal{AC} = \hat{y}^1, ..., \hat{y}^n$.

\textbf{Argument Relation Identification (ARI)}: In this task, the input is an essay $D_c$ segmented into $m$ discourse chunks, i.e., $D_c = [c^1, ..., c^m]$. Given the argument component type of each chunks, the model is required to identify and classify all possible argument relations between discourse chunks $\mathcal{AR} = \{(i_{from}, i_{to}, r)\}, i_{from},i_{to}\in[1,m]$. It is worth noting that relational instances are sparse within individual essays, leading to a high ratio of negative to positive samples, which make this task more complexed.

The three tasks are arranged in ascending order of difficulty. The AES task is rather simple, while the ACD task necessitates more domain knowledge though manageable and relatively easy in general. The ARI task, however, combines complexity and domain-specific expertise, posing a significant challenge to the capabilities of LLMs.

\subsection{Dataset}

\begin{itemize}
    \item \textbf{AEE} \cite{stab-gurevych-2017-parsing} was annotated on student-written essays. A stance for a controversial theme is expressed by a major claim component as well as claim components, and premise components justify or refute the claims. Attack and support labels are defined as relations. 
    \item \textbf{\DatasetName}~\cite{ren2025comprehensiveargumentanalysiseducation} also contains argumentative essays penned by students. It defines 10 fine-grained categories of argument components and 14 types of argument relations, posing additional challenges in the comprehension and evaluation of argumentative essays. Moreover, in \DatasetName, a single sequence pair may be associated with multiple relation types, which adds additional challenges to the task. 
    \item \textbf{ASAP 2.0} \cite{CROSSLEY2025100954} is a representative dataset in essay scoring, which collected a large amount of student-written argumentative essays. Given its large scale, we sampled 1000 essays to conduct experiments.
    \item \textbf{ArGPT} \cite{Rocha2024AssessingGB} includes arguments generated by ChatGPT and annotated labels assessing the quality. We incorporate this dataset into our experiments with the aim of providing insights for applications involving LLM-based scoring or evaluation.
\end{itemize}

Additional details about these dataset are provided in Appendix \ref{adx:dataset}.

\subsection{Baselines and Evaluation metrics}

As baselines, we employ DeepSeek-R1 as base model (for details, see Appendix \ref{sec:implementation}) and report the performance of several existing methods, including In-Context Learning (ICL), Chain of Thought (CoT), and Calibrate \cite{Liu2023CalibratingLE}, which an prompt optimization framework that follows a three-stage framework consisting of drafting, filtering, and refining to generate final criteria. 

For the AES task, we additionally compare our approach with the methods proposed by \citet{Stahl2024ExploringLP}, who investigated various prompt strategies for automated scoring of argumentative essays. Specifically, we select the Feedback\_dCoT$\rightarrow$Score and Explanation$\rightarrow$Score strategies as baselines due to their relatively strong performance as reported in the original study. Different subtasks are evaluated using task-specific metrics:

\begin{enumerate}
    \item \textbf{AES:} For this scoring task, we use Quadratic Weighted Kappa (QWK) as the primary evaluation metric, providing a nuanced measure of inter-rater reliability by considering the relative difference between scores. 
    \item \textbf{ACD and ARI:} For these classification tasks, we report both Micro F1 and Macro F1 scores (denoted as \textit{Micro} and \textit{Macro} in tables) to provide a comprehensive evaluation.
\end{enumerate}

\section{Main Results}
\label{sec:results}

The main results of our experiments is shown in Table \ref{tab:main_results} and Table \ref{tab:main_results_scoring}, where we report the performance of our framework in all three subtasks on different datasets.

\begin{figure}[h]
    \centering
    \begin{tikzpicture}
        \begin{axis}[
            ylabel={Error},
            xlabel={Iteration}, 
            xmin=0, xmax=240, 
            ymin=0, ymax=0.6, 
            xtick={0,40,80,120,160,200,240}, 
            ytick={0,0.1,0.2,0.3,0.4,0.5,0.6}, 
            width=0.45\textwidth,
            grid=both, 
            legend pos=south east 
        ]
        \addplot[
            color=red, 
            mark=*, 
        ]
        coordinates {
            (0,0.375)(40,0.525)(80,0.425)(120,0.4)(160,0.375)(200,0.425)(240,0.375)
        };
        \addlegendentry{Criteria-based} 

        \addplot[
            color=blue, 
            mark=*, 
        ]
        coordinates {
            (0,0.375)(40,0.425)(80,0.3)(120,0.325)(160,0.3)(200,0.3)(240,0.325)
        };
        \addlegendentry{\FrameworkNameShort} 
        \end{axis}
    \end{tikzpicture}
    \caption{The error dynamic during optimizing process of AES, where error is computed via the absolute difference between predicted and labeled scores.}
    \label{fig:discrepancy_comparison}
\end{figure}
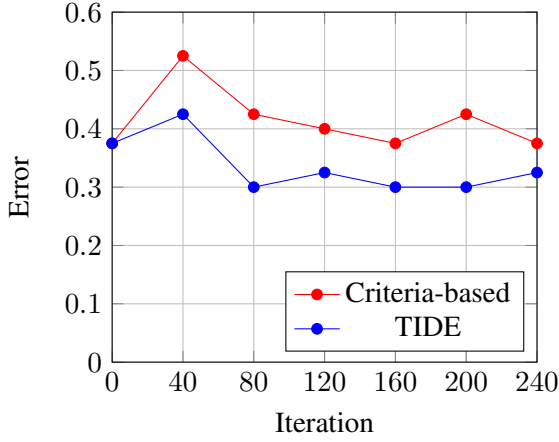

In the AES task, the Criteria-based approach yielded relatively high performance in terms of QWK. This pattern suggests that \MetaSolution~is effective at capturing the relative ranking or category of the scores. In contrast, \FrameworkName~achieved a higher score of QWK, which demonstrates more consistent and robust performance, exhibiting lower sensitivity to individual samples that could otherwise compromise overall model effectiveness. As illustrated in Figure \ref{fig:discrepancy_comparison}, the reduced error magnitude further substantiates the efficacy of our proposed design.

\begin{table}[htb]
    \setlength{\arraycolsep}{1pt} 
    \centering
    \begin{tabular}{l|c|c|c}
        \toprule
        Method & \DatasetName & ArGPT & ASAP 2.0\\
        \midrule
        ICL & 25.49 & 13.60 & 25.10\\
        CoT & 16.08 & 13.22 & 25.99\\
        Calibrate  & 10.81 & 5.04 & 16.69\\
        Feed & 10.20 & 15.73 & 20.00\\
        Exp & 15.33 &34.34 & 15.08\\
        \hline
        Criteria & 24.17 & 48.25 & 38.61 \\
        \FrameworkNameShort & \textbf{45.64} & \textbf{53.59} & \textbf{49.51}\\
        \bottomrule
    \end{tabular}
    \caption{Performance comparison with different baselines on AES. In this table 'Feed' and 'Exp' represents Feedback\_dCoT and Explanation from \citet{Stahl2024ExploringLP}, while 'Criteria' represents the criteria-based prompt optimizing. All results in this table are presented in percent format(\%).}
    \label{tab:main_results_scoring}
\end{table}

For the ACD task, the Criteria-based method attained the second highest Micro F1 score but performed worst in terms of Macro F1. This discrepancy indicates that \MetaSolution~is particularly discrepancyed towards frequent labels. Conversely, \FrameworkName~maintains balanced performance, demonstrating the ability to correctly classify both frequent and rare labels, thus enhancing overall robustness in label distribution.

\begin{table}[htbp]
    \centering
    \begin{tabular}{l|c|c|c}
         \hline
         Method & Precision & Recall & F1  \\
         \hline
         ICL & 27.82 & 16.38 & 20.62 \\
         CoT & 30.82 & 18.38 & 23.03 \\
         Calibrate & \underline{40.49} & \underline{26.71} & \underline{32.19} \\
         \hline
         Criteria-based & 29.08 & 18.19 & 22.38 \\
         \FrameworkNameShort & \textbf{40.80} & \textbf{26.78} & \textbf{32.33} \\
         \hline
    \end{tabular}
    \caption{Predictions of whether there is some relation between two chunks of discourse in task ARI from \DatasetName. All results in this table are presented in percent format(\%).}
    \label{tab:ari_hit}
\end{table}

\begin{figure*}[htbp]
    \centering
    \begin{subfigure}[b]{0.45\textwidth}
        \includegraphics[width=\textwidth]{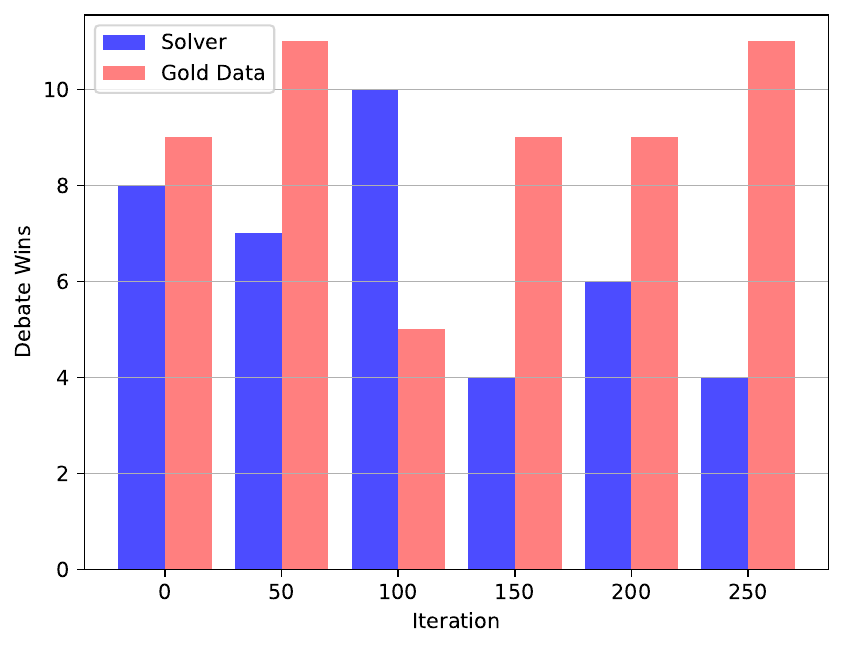}
        \caption{\RetryTimesMark=2}
        \label{fig:win_rate_retry2}
    \end{subfigure}
    \hfill
    \begin{subfigure}[b]{0.45\textwidth}
        \includegraphics[width=\textwidth]{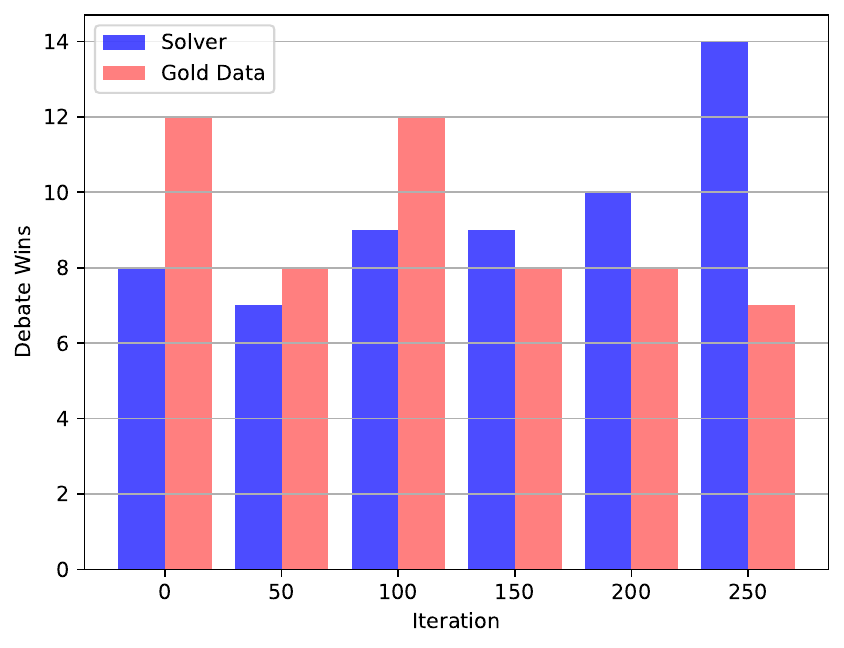}
        \caption{\RetryTimesMark=4}
        \label{fig:win_rate_retry4}
    \end{subfigure}
    \caption{Debate Wins for AES on \DatasetName~in different settings}
    \label{fig:win_rate_combined}
\end{figure*}

In the ARI task, \FrameworkName~surpassed all other methods by achieving superior performance in both Micro and Macro F1 scores. To gain deeper insights, we analyzed whether the models were capable of correctly identifying the presence of relations between chunks, i.e. telling the positive samples from negative ones. As presented in Table \ref{tab:ari_hit}, the Criteria-based approach improved the detection of discourse relations relative to the ICL baseline, and also enhanced relation-type classification (Table \ref{tab:main_results}). Notably, although Calibrate and \FrameworkNameShort~exhibited comparable performance in general metrics, \FrameworkName~significantly outperformed Calibrate on ARI evaluations. These results further underscore the effectiveness and generalizability of our proposed framework. The output criteria can be found in Appendix \ref{adx:opt_samples}.

\subsection{Reasoning Models Leads to Better Performance}
\label{sec:compare_models}

Table~\ref{tab:compare_model} presents an evaluation of different models employed as \MetaSolution~and \Solution. Our primary focus is on assessing the impact of the reasoning capability of DeepSeek-R1, which proves to be essential for generating and optimizing criteria as shown in the table. 

\begingroup
\setlength{\tabcolsep}{2pt}
\begin{table}[h]
    \centering
    \begin{tabular}{l|c|cc|cc}
        \toprule
        \multirow{2}{*}{Role Settings} & \multicolumn{1}{c|}{AES} & \multicolumn{2}{c|}{ACD}& \multicolumn{2}{c}{ARI}\\
        & QWK & Micro & Macro & Micro & Macro\\
        \midrule
        R1+R1 & \textbf{45.64} & \textbf{69.57} & \textbf{56.85} & \textbf{16.05} & \textbf{15.39} \\
        R1+V3 & \underline{45.03} & \underline{66.72} & \underline{48.16} & \underline{11.63} & \underline{12.66} \\
        V3+R1 & 44.94 & 64.77 & 44.24 & 6.38 & 7.11 \\
        \hline
        o1-mini & 11.56 & 59.98 & 35.23 & 7.53 & 7.18 \\
        QwQ & 17.19 & 27.78 & 42.04 & 7.32 & 7.18 \\
        \hline
        Qwen-TIDE & 10.62 & 59.88 & 40.16 & 6.17 & 5.83 \\
        Qwen-R1 & 29.19 & 63.08 & 45.47 & 10.15 & 8.75 \\
        \bottomrule
    \end{tabular}
    \caption{Different combination of (\MetaSolution+\Solution). For example, R1+V3 represents employing DeepSeek-R1 as \MetaSolution, while DeepSeek-V3 as \Solution. QwQ+QwQ represents employ QwQ-32B for both \MetaSolution~and \Solution. All results in this table are presented in percent format(\%).}
    \label{tab:compare_model}
\end{table}
\endgroup

The results demonstrate that utilizing DeepSeek-R1 as both \MetaSolution~and \Solution~leads to significant performance improvements across all tasks. More specifically, when the reasoning ability is removed from \MetaSolution~(i.e., the V3+R1 setting in Table~\ref{tab:compare_model}), \FrameworkName~suffers a substantial decline in performance across AES, ACD, and ARI. This finding highlights the critical role of a reasoning-capable \MetaSolution~in driving the framework’s success. Furthermore, although the R1+V3 setting performs better than V3+R1, it still underperforms compared to the R1+R1 configuration. This indicates that a strong \Solution~is also necessary to fully leverage the benefits of the framework. We also included the results provided by o1-mini and QwQ in the table, where DeepSeek-R1 consistently outperforms o1-mini and QwQ across all tasks. This is possibly due to their pretraining focus on mathematical and code reasoning tasks instead of arguments \cite{ElKishky2024OpenAIOS,qwq32b}.

In addition, to further study the generalization of \FrameworkNameShort, we tested a small scaled model, Qwen2.5-14B using criteria derived from Deepseek-R1’s TIDE runs (Qwen-R1 row in the table). The results show clear performance improvements across all three tasks compared to the native Qwen-TIDE setup, even outperforming o1-mini and QwQ, and coming reasonably close to R1+V3 in Table \ref{tab:compare_model}. These findings demonstrate the generalization capability of TIDE, as well as reducing costs on influence phase. 

\subsection{Complex Tasks need More Trial}
\label{sec:compare_settings}

In this section we discuss different settings of \FrameworkName, which mainly includes \RetryMark~and batch size, which is presented in Table \ref{tab:compare_settings}. It is evident that, for the ARI task—which is comparatively more complex and requires a higher degree of domain knowledge—a larger number of \RetryMark~is more suitable. This is because \MetaSolution~benefits from multiple refinement steps to achieve more effective model updates. The impact of \RetryMark~is further accentuated when considering the effect of batch size: smaller batch sizes consistently yield better performance, likely due to improved granularity in learning signal and reduced averaging effects. In contrast, for the AES task—which is relatively simpler in nature—increasing the number of \RetryMark~or using smaller batch sizes tends to degrade performance. Specifically, such configurations result in reduced QWK scores, thereby mimicking the shortcomings observed in the Criteria-based approach, as discussed in Table \ref{tab:main_results}. As for ACD, which lies between AES and ARI in terms of task complexity, optimal performance is achieved with moderate  settings. This suggests that both the number of \RetryMark~iterations and batch size should be chosen to reflect the intermediate difficulty of the task, balancing the trade-off between refinement and overfitting.

\begingroup
\setlength{\tabcolsep}{4pt}
\begin{table}[h]
    \centering
    \begin{tabular}{l|c|cc|cc}
        \toprule
        \multirow{2}{*}{Settings} & \multicolumn{1}{c|}{AES} & \multicolumn{2}{c|}{ACD}& \multicolumn{2}{c}{ARI}\\
        & QWK & Micro & Macro & Micro & Macro\\
        \midrule
        \RetryTimesMark=1 & 38.14 & 67.29 & 40.97 & 10.55 & 11.49 \\
        \RetryTimesMark=2 & \textbf{45.64} & \textbf{69.57} & \textbf{56.85} & 12.64 & 13.16 \\
        \RetryTimesMark=4 & 21.85 & 68.92 & 51.33 & \textbf{16.05} & \textbf{15.39} \\
        \hline
        $bsz$=1 & 19.66 & 64.80 & 45.93 & \textbf{14.07} & \textbf{13.64} \\
        $bsz$=2 & \textbf{45.64} & \textbf{69.57} & \textbf{56.85} & 12.64 & 13.16 \\
        $bsz$=4 & 40.00 & 68.29 & 54.71 & 9.44 & 10.01 \\
        \bottomrule
    \end{tabular}
    \caption{Different setting of \RetryMark~and batch in \FrameworkName. When discussing \RetryMark, we set batch size to 2. When discussing batch, we set \RetryMark~to 2. All results in this table are presented in percent format(\%).}
    \label{tab:compare_settings}
\end{table}
\endgroup

\subsection{Simple Tasks need More Debate}

The results in Table \ref{tab:abl_debate} clearly demonstrate that the incorporation of the \DebateMark~mechanism significantly improves both Pearson and QWK scores, which indicates its effectiveness in reducing noise and enabling more accurate and stable predictions during training.


\begin{table}[htbp]
    \centering
    \begin{tabular}{c|cc}
    \hline
    & Pearson & \textbf{QWK}  \\
    \hline
    \multicolumn{3}{l}{\RetryTimesMark=1} \\
    \hline
         w/o debate & 25.59 & 24.17 \\
         w/ debate & \textbf{42.87}(+17.28) & \textbf{38.14}(+13.97) \\
    \hline
    \multicolumn{3}{l}{\RetryTimesMark=2} \\
    \hline
         w/o debate & 42.04 & 38.31 \\
         w/ debate & \textbf{52.59}(+10.55) & \textbf{45.64}(+7.33) \\
    \hline
    \end{tabular}
    \caption{Ablation study of \DebateMark~in AES on \DatasetName. We computed another metric: Pearson coefficient, in order to provide more insights.}
    \label{tab:abl_debate}
\end{table}

The win rate of \Solution~in debate during AES execution is illustrated in Figure \ref{fig:win_rate_combined}. Notably, under the \RetryTimesMark=4 setting, \Solution~achieves a substantially higher win rate compared to the Gold Data. This may, to some extent, hinder the system's ability to continuously improve through training. In contrast, with \RetryTimesMark=2, \Solution~exhibits a considerably lower win rate, suggesting that the system continues to optimize, which corresponds to the superior performance reported in Table \ref{tab:main_results}. Further discussion and details on the win rate are provided in Appendix \ref{adx:win_rate}.

\section{Conclusion}
In this work, we propose \FrameworkName, a novel framework designed to enhance performance of understanding and evaluating argumentative essays through a combination of iterative refinement and structured interaction between two main roles—\MetaSolution~and \Solution. Through studies, we show that key components such as the \RetryMark~mechanism and the \DebateMark~module play pivotal roles in balancing robustness and performance. Specifically, we find that task complexity should guide the configuration of these components: while simpler tasks like AES benefit from minimal refinement but need more debate, while more complex tasks like ARI require deeper iterative processing to achieve optimal outcomes.

Moreover, our analysis reveals that, for DeepSeek series models, the reasoning ability of the underlying models is crucial: both the \MetaSolution~and the \Solution~must possess strong reasoning capabilities to fully realize the potential of \FrameworkName. In addition, other weak, small-scaled models can also benefit from the outcome criteria from these powerful ones, which shows the generalizability of the framework.

\section*{Limitations}

While \FrameworkName~demonstrates promising performance across a range of educational NLP tasks, several limitations remain that warrant further exploration.

\begin{itemize}
    \item \textbf{Debate Protocol Variants.} In the current implementation of \FrameworkNameShort, we adopt a conventional debate protocol in which two parties represent opposing stances and take turns to defend their respective positions. Prior work \citep{Khan2024DebatingWM} has proposed alternative debate paradigms, including consultancy-style discussions, structured debates, and interactive debates that allow for dynamic exchanges. Exploring how these different debate protocols influence the quality and stability of prompt optimization within \FrameworkNameShort~may lead to further performance gains and a deeper understanding of model reasoning dynamics.
    \item \textbf{Cross-domain Generalizability.} Although our experiments demonstrate the effectiveness of \FrameworkName~across three argumentation-related tasks in the educational domain, its applicability to other fields remains uncertain. Domains such as law, medicine, and policy analysis present unique linguistic structures, reasoning requirements, and data distributions. Investigating the adaptability of \FrameworkName~to these high-stakes or domain-specific applications—particularly in terms of prompt robustness, criteria abstraction, and reasoning fidelity—constitutes an important direction for future research.
\end{itemize}


\appendix



\section{Implementation Details}
\label{sec:implementation}

In this work, we mainly employ Deepseek-R1 as the backbone model, due to its powerful capability shown in long CoT reasoning and low in cost. Also, we compare it with o1-mini \cite{Contributors2024OpenAIOS}, QwQ \cite{qwq32b}, a small model Qwen2.5-14B \cite{Yang2024Qwen25TR}, as well as Deepseek V3 \cite{DeepSeekAI2024DeepSeekV3TR}, which is the base model of Deepseek R1 but do not feature in o1-like reasoning, in order to investigate the influence of reasoning ability in our framework. 

To better guide \MetaSolution~in generating more aligned initial criteria, we incorporate several in-context demonstrations during this stage. In addition, unless explicitly stated, the batch size is set to 2 in our experiments. Furthermore, to minimize the influence from the \Solution~model, we set its temperature to 0.7.

For the baselines, we include two demonstrations in the context of both ICL and CoT. To mitigate prompt-induced discrepancy, we adopt the same prompt template used for \Solution~when generating predictions in ICL, and append “Let’s think step by step...” for CoT. Similarly, for Calibrate, we follow the same setup, but replace the updating criteria with the four atomic editing operations proposed in the original paper.

\section{Token Budgets}

Given the iterative nature of TIDE, it inherently entails a relatively high token budget. This aspect constitutes one of the primary motivations for selecting DeepSeek-R1 as the base model for TIDE, owing to its favorable cost-efficiency and competitive performance. The token consumption of both Criteria-based and TIDE methods during the prompt optimization process is documented in Table \ref{tab:token_consumption}.

\begin{table}[h]
    \centering
    \begin{tabular}{c|c|c|c}
        \hline
         Method & AES & ACD & ARI \\
         \hline
         Criteria &  2.27M & 3.38M & 5.51M \\
         TIDE & 1.08M & 4.46M & 13.27M \\
         \hline
    \end{tabular}
    \caption{Token budget for Criteria-based method and \FrameworkNameShort, where DeepSeek-R1 was adopted as the backbone model.}
    \label{tab:token_consumption}
\end{table}

The introduction of the Debate and Trial modules intuitively increases TIDE's token consumption compared to the Criteria-based method, as the trade-off of the performance. However, as shown in the table, we observe the opposite trend on AES. One possible reason is that the Guider achieves a certain number of victories during the Debate module, thus reducing the number of updates. 

We also incorporate insights aimed at reducing the overall cost, as discussed in Section~\ref{sec:compare_models}. During inference, an alternative approach is to employ lightweight models rather than powerful but costly ones, leveraging the optimized criteria to predict labels. Although this substitution may incur a slight performance degradation, it can significantly reduce API invocation costs or GPU resource consumption.


\section{Debate Win Rate}
\label{adx:win_rate}

To further dive into the win rate of \Solution, we examine the performance at the iteration point where \Solution’s win rate surpasses that of the Gold Data under \RetryMark=4 on Iteration 180. The results, presented in Table \ref{tab:win_rate_compare}, reveal a noticeable performance jump—particularly in Pearson and QWK metrics—between iteration 180 and iteration 300. Further investigation into controlling \Solution’s win rate remains an open direction for future work.

Additionally, we observe that \Solution~rarely wins debates during training in ACD and ARI. For instance, under the batch size of 2, only five wins are observed in ACD and two in ARI over 240 iterations. We argue that this takes place when the task is relatively complex, compared the scoring task where the model itself is already capable (though not aligned with the data).

\begin{table}[htbp]
    \centering
    \begin{tabular}{c|cccc}
    \hline 
        Settings & Pearson & QWK \\
    \hline 
        $Iter=180$ & \textbf{30.25} & \textbf{29.21}  \\
        $Iter=300$ & 22.11 & 21.85  \\
    \hline
    \end{tabular}
    \caption{The performance in different settings in AES task, where \RetryMark=4 and batch size is set to 2.}
    \label{tab:win_rate_compare}
\end{table}


\section{Discrepancy Computation}
\label{adx:discrepancy}

For different task, we employed different methods for computing discrepancy, all in document-level. For AES, we utilize the absolute difference between $y$ and $\hat{y}$ as the discrepancy, while number of labels that mismatch with $y$ for each discourse in ACD.

$$d_{AES} = abs(y - \hat{y})$$

$$d_{ACD} = |\{\hat{y}_i|\hat{y}_i \neq y_i\}|$$

For ARI, in addition to mismatched labels between individual pairs of chunks (i.e., predicted labels not appearing in the ground truth $y$, and ground truth labels not present in the predictions $\hat{y}$), we also consider the accuracy of pairwise identification. To further enhance the precision of index extraction, we impose penalty on cases with incorrect index predictions, as illustrated in Algorithm \ref{alg:ari_discrepancy}. Specifically, we count the number of predicted pairs that do not exist in $y$, as well as the number of ground truth pairs in $y$ that are not predicted. These errors are penalized with a weight of 2 in our experiments. This design encourages \FrameworkNameShort~to more accurately identify chunk pairs with relational links, as demonstrated in Table~\ref{tab:ari_hit}.

\begin{figure}[h]
    \centering
    \begin{tikzpicture}
    \begin{axis}[
        xlabel={Iteration},
        ylabel={Error},
        xmin=0, xmax=240,
        ymin=2, ymax=4.6,
        xtick={0,30,...,300},
        ytick={2,2.4,...,4.6},
        grid=both,
        width=0.45\textwidth,
        legend pos=south east 
    ]
    \addplot [
        color=red,
        mark=square,
        ]
        coordinates {
        (0, 3.8666666666666667)
        (30, 4.5)
        (60, 3.5)
        (90, 3.933333333333333)
        (120, 3.566666666666667)
        (150, 3.8666666666666667)
        (180, 4.433333333333334)
        (210, 4.033333333333333)
        (240, 3.5)
        };
        \addlegendentry{Criteria-based} 
    
    \addplot[
        color=blue,
        mark=square,
        ]
        coordinates {
        (0,3.2333333333333334)
        (30,3.466666666666667)
        (60,2.8666666666666667)
        (90,3.3)
        (120,3.6666666666666665)
        (150,3.1333333333333333)
        (180,2.8333333333333335)
        (210,3.1333333333333333)
        (240,3.066666666666667)
        (270,3.3)
        };
        \addlegendentry{\FrameworkNameShort} 
        
    \end{axis}
    \end{tikzpicture}
    \caption{Error dynamic during training for ACD}
    \label{fig:discrepancy_acd}
\end{figure}
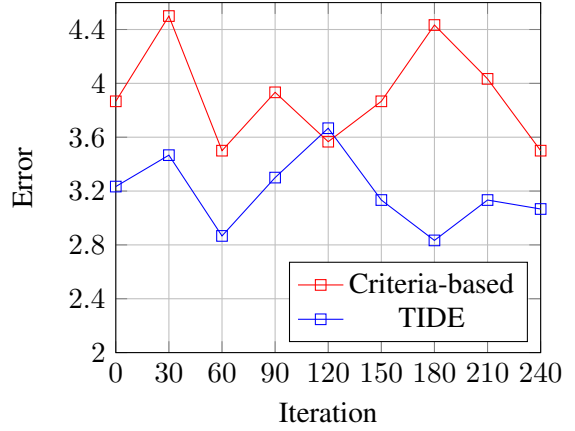

\begin{figure}[h]
    \centering
    \begin{tikzpicture}
    \begin{axis}[
        xlabel={Iteration},
        ylabel={Error},
        xmin=0, xmax=240,
        ymin=20, ymax=32,
        xtick={0,30,...,240},
        ytick={20,22,...,32},
        grid=both,
        width=0.45\textwidth,
        legend pos=south east 
    ]
    
    \addplot[
        color=red,
        mark=square,
        ]
        coordinates {
        (0, 24.533333333333335)
        (30, 27.633333333333333)
        (60, 24.266666666666666)
        (90, 23.866666666666667)
        (120, 23.9)
        (150, 31.066666666666666)
        (180, 26.333333333333332)
        (210, 24.433333333333334)
        (240, 26.1)
        (270, 24.666666666666668)
    };
    \addlegendentry{Criteria-based} 
    
    \addplot[
        color=blue,
        mark=square,
        ]
        coordinates {
        (0,23.733333333333334)
        (30,25.233333333333334)
        (60,26.066666666666666)
        (90,28.933333333333334)
        (120,29.033333333333335)
        (150,25.833333333333332)
        (180,27.266666666666666)
        (210,24.233333333333334)
        (240,23.133333333333333)
        (270,22.733333333333334)
        };
    \addlegendentry{\FrameworkNameShort} 
    \end{axis}
    \end{tikzpicture}
    \caption{Error dynamic during training for ARI}
    \label{fig:discrepancy_ari}
\end{figure}
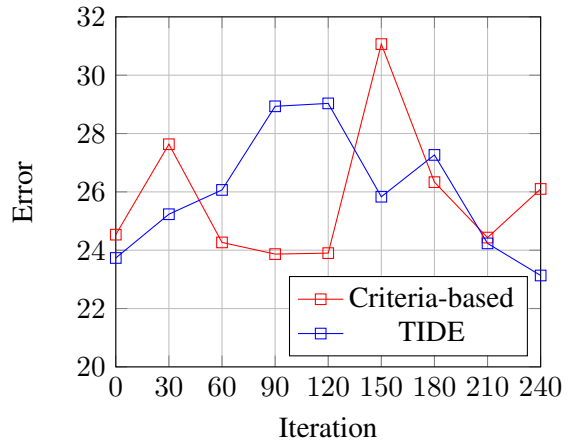

\begin{figure}[htbp] 
\centering
\begin{algorithm}[H] 
\caption{Computation for $Discrepancy_{ari}$}
\label{alg:ari_discrepancy}
\begin{algorithmic}[1]
    \Require Ground truth pairs $Y$, predicted pairs $\hat{Y}$, penalty $p$
    \Ensure Discrepancy for $\hat{Y}$
    
    \State $d_{ARI} \gets 0$
    
    \For{$\hat{y} \in \hat{Y}$}
        \State Ground truth pair that match in index $y, y \in Y~and~y_{from}=\hat{y}_{from} ~and~y_{to}=\hat{y}_{to}$
        \If{y is not null}
            \State $d_{ARI} \gets d_{ARI} + $ mismatched labels between $y$ and $\hat{y}$
        \Else
            \State $d_{ARI} \gets d_{ARI} + $ number of labels in $\hat{y} + 2 \times p$
        \EndIf
    \EndFor
    
    \For{$y \in Y$}
        \If{$y$ not be predicted in previous loop}
            \State $d_{ARI} \gets d_{ARI} + $ number of labels in $y + 2 \times p$
        \EndIf
    \EndFor
    
    \State \Return $d_{ARI}$
\end{algorithmic}
\end{algorithm}
\end{figure}

The discrepancy dynamic during training for AES is presented in Figure \ref{fig:discrepancy_comparison}, while for ACD and ARI is illustrated in Figure \ref{fig:discrepancy_acd} and Figure \ref{fig:discrepancy_ari}.


\section{Details of Dataset}
\label{adx:dataset}

In this work, we utilize \DatasetName~, ArGPT, AEE and ASAP 2.0 to evaluate the performance of \FrameworkNameShort. For ASAP 2.0, given its large scale, we randomly shuffle and select 10\% samples to conduct experiments. For AEE, we follow the original train-test split from the original paper. For the other datasets, we randomly shuffle the data using a fixed seed of 42, and then split each dataset into 60\% for training and 40\% for evaluation to ensure sufficient evaluation samples.

\subsection{\DatasetName}
\DatasetName~\cite{ren2025comprehensiveargumentanalysiseducation} includes 226 Chinese argumentative essays penned by high school students. These essays range from 557 to 1,101 tokens with an average of 829.82 tokens. There are 4,726 dicourse in total, each of which has an argument component category in \textit{MajorClaim, Claim, Restated Claim, Fact, Anecdote, Quotation, Proverb, Axiom, Elaboration} and \textit{Others}, which is what the label needs to be predicted in ACD task. For ARI, the dataset defines 14 fine-grained categories of \textit{Positive, Negative, Comparative, Example, Citation, Metaphorical, Hypothetical, Restatement, Detail, Background, Coherence, Progression, Contrast} and \textit{Concession}. There are 4,837 relations appear in the chunks. Following \citet{Ren2024CEAMCCA}, we categorize the original score data into five levels, corresponding to scores from 1 to 5.

The original paper mainly reported performance of pretrained models such as RoBERTa \cite{Liu2019RoBERTaAR} and Longformer \cite{Beltagy2020LongformerTL}, and LLMs after fine-tuning such as ChatGLM \cite{Zeng2024ChatGLMAF} and Qwen \cite{Bai2023QwenTR}. In this work we mainly discuss how \FrameworkNameShort~help improve the performance via prompt optimization.

\subsection{AEE}

The AEE dataset focuses on analyzing argumentative essays by annotating the argument components and relations within. The dataset contains over 400 student-written argumentative essays. We include this dataset mainly because its division of components (Major Claim, Claim and Premise) and relations (Attack and Support) is more close to conventional argument mining.

\subsection{ASAP 2.0}

Inherited from the representative dataset for automatic essay scoring ASAP, a newly-updated ASAP 2.0 dataset was promoted. This dataset incorporate about 24,000 student-written argumentative essays, aligned to the latest standards for student-appropriate assessments. It also included samples across economic and location populations to mitigate the potential of algorithmic bias. 

\subsection{ArGPT}

The ArGPT dataset primarily targets the evaluation of argument quality in texts generated by ChatGPT. It consists of 168 argumentative essays, carefully constructed through simulated student-professor dialogues to elicit diverse argument structures. We included this dataset with the aim of providing insights for applications that utilize LLM-based evaluation to give a score. However, in comparison to \DatasetName, the definitions of argument components (Major Claim and Premise) and their relations (Attack and Support) are relatively simplistic. Therefore, we primarily utilize the AES task from this dataset to evaluate the effectiveness of \FrameworkNameShort.

\begin{figure}[h]
    \centering
    \begin{tikzpicture}
        \begin{axis}[
            xlabel={Iteration},
            ylabel={Length},
            xmin=0, xmax=240,
            ymin=400, ymax=800,
            xtick={0,40,...,300},
            ytick={400,450,...,800},
            grid=both,
            width=0.45\textwidth
        ]
        
        \addplot[
            color=blue,
            mark=square
        ]
        coordinates {
            (0,405)
            (20,540)
            (40,696)
            (60,714)
            (80,600)
            (100,664)
            (120,664)
            (140,622)
            (160,528)
            (180,498)
            (200,577)
            (220,600)
            (240,695)
        };
    \end{axis}
    \end{tikzpicture}
    \caption{Length dynamic during iteration for AES}
    \label{fig:length_aes}
\end{figure}
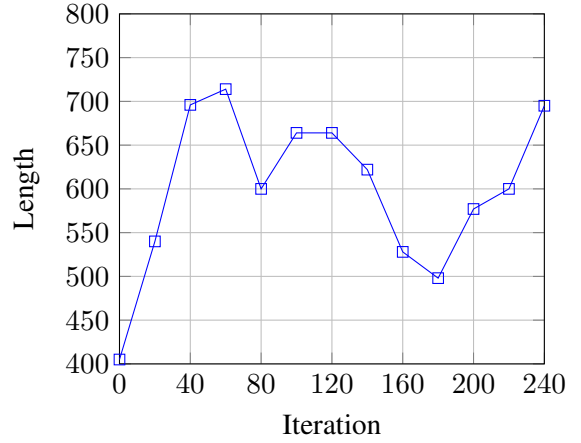

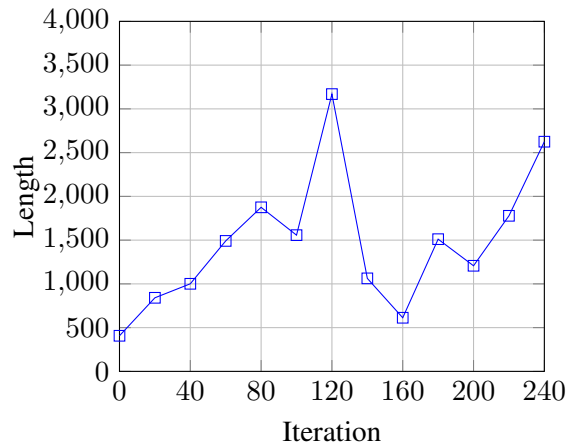
\begin{figure}[h]
    \centering
    \begin{tikzpicture}
    \begin{axis}[
        xlabel={Iteration},
        ylabel={Length},
        xmin=0, xmax=240,
        ymin=0, ymax=4000, 
        xtick={0,40,...,300},
        ytick={0,500,...,4000},
        grid=both,
        width=0.45\textwidth
    ]
    
    \addplot[
        color=blue,
        mark=square
    ]
    coordinates {
        (0,407)
        (20,841)
        (40,1001)
        (60,1491)
        (80,1875)
        (100,1557)
        (120,3169)
        (140,1063)
        (160,613)
        (180,1511)
        (200,1207)
        (220,1778)
        (240,2625)
    };
    
    \end{axis}
    \end{tikzpicture}
    \caption{Length dynamic during iteration for ACD}
    \label{fig:length_acd}
\end{figure}

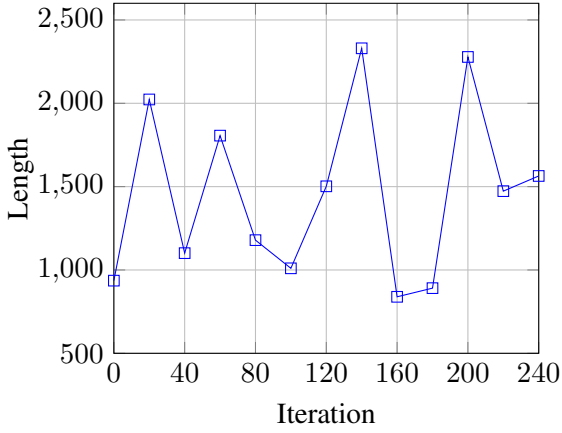
\begin{figure}[h]
    \centering
    \begin{tikzpicture}
    \begin{axis}[
        xlabel={Iteration},
        ylabel={Length},
        xmin=0, xmax=240,
        ymin=500, ymax=2600, 
        xtick={0,40,...,300},
        width=0.45\textwidth,
        ytick={0,500,...,3500},
        grid=both
    ]
    
    \addplot[
        color=blue,
        mark=square
    ]
    coordinates {
        (0,936)
        (20,2023)
        (40,1101)
        (60,1806)
        (80,1179)
        (100,1010)
        (120,1502)
        (140,2330)
        (160,839)
        (180,891)
        (200,2278)
        (220,1473)
        (240,1564)
    };
    
    \end{axis}
    \end{tikzpicture}
    \caption{Length dynamic during iteration for ARI}
    \label{fig:length_ari}
\end{figure}




\section{Output Samples}
\label{adx:opt_samples}

In this section we present the final output from \FrameworkNameShort~in Table \ref{tab:aes_result}, \ref{tab:acd_result} and \ref{tab:ari_result}, respectively. It is obvious that after iterations of refinement, the length of criteria extends with more details of each category included compared to the original one, which is mainly caused by learning different features through training, as shown in Figure \ref{fig:length_aes}, Figure \ref{fig:length_acd} and Figure \ref{fig:length_ari}. In addition, as shown in Table \ref{tab:ari_result}, \MetaSolution~learns to develop a quantification principle during the iterative process in ARI, which enables the model to better adapt to the task. This is further demonstrated in Table \ref{tab:case_ari}. This capability allows \Solution~to more effectively distinguish between different categories, thereby enhancing overall performance in ARI.

\begin{table*}[]
    \centering
    \begin{tabularx}{\textwidth}{X}
        \hline 
        Before refinement \\
        \hline 
        The determination of argumentative relationship categories is based on semantic functions and logical structures,combined with the categorical properties of argumentative components.The characteristics are as follows: Example: Using specific examples(historical facts/personal instances)to directly support the thesis,forming a concrete→abstract supporting relationship between the example and the thesis,often containing empirical elements. $...$ \\
        \hline  
        After refinement \\
        \hline 
        The determination of argumentative relationship categories is based on semantic functions and logical structures, combined with the category attributes of argumentative components. The characteristics are as follows:1. Example-based argumentation: Historical examples must fully present the causal chain of motivation-action-impact (for cross-sentence combinations, stage completeness markers + \textcolor{blue}{logical deduction words $\geq$ 5} are required). New exclusion conditions: e) Case enumeration that does not directly correspond to the core elements of the sub-argument (\textcolor{blue}{with an element mapping degree < 85\% and deduction words < 5}). f) Quotations that do not form a complete argumentative chain (must meet the \textcolor{blue}{sub-argument element mapping degree $\geq$ 90\% and have deduction words $\geq$ 5}). g) Historical examples that do not fully present any stage of motivation, action, or impact. $...$ \\
        \hline 
    \end{tabularx}
    \caption{Case study of ARI, where \MetaSolution~learned to develop a principle of quantification during iteration.}
    \label{tab:case_ari} 
\end{table*}

\begin{table*}[htb]
    \centering
    \begin{tabularx}{\textwidth}{X}
        \hline
            Five-Dimensional and Five-Level Scoring System for High School Argumentative Essays(1-5 Points) - Innovation of Thesis(Weight:8\%) - 5 Points: Dual-dimension comparison including cultural symbol comparison(cross-regional/generational)+implicit opposition$\geq$2 groups(including at least 1 group of philosophical/cultural conflict)+at least 1 example from after 2015(philosophical examples may be used as substitutes). - 4 Points: Single-dimension analysis with implicit opposition$\geq$1 group(clearly specifying the type of opposition)+allowing classical philosophical examples to replace modern examples. - 3 Points: Single-case argumentation+implicit opposition$\geq$1 group(philosophical attributes must be clearly specified). - Effectiveness of Argumentation(Weight:38\%) - 5 Points: Three-tiered progressive structure(phenomenon-essence-value)+cross-temporal and cross-spatial case corroboration$\geq$2 groups(spanning at least 2 different fields or time periods)+argumentation layers$\geq$4(including at least 1 layer of counter-proof). - 4 Points: Two-tiered progression+cross-era case comparison(time span$\geq$5 years)+argumentation layers$\geq$3(supported by data or philosophical reasoning). - Quality of Evidence(Weight:20\%) - 5 Points: Case analysis$\geq$100 words(including deconstruction of contradictions)+positive-to-negative ratio$\geq$1.2:1+at least 1 citation from academic literature or philosophical classics. - 4 Points: Case analysis$\geq$80 words(including deconstruction of symbols)+positive-to-negative ratio$\geq$1:1+allowing philosophical examples to replace literature citations. - Precision of Expression(Weight:14\%) - 5 Points: Composite rhetorical density$\geq$0.7 types per 100 words(including at least 1 type of rhetorical nesting)+logical density$\geq$7\%(including at least 3 types of logical connectors). - 4 Points: Composite rhetorical density$\geq$0.5 types per 100 words+logical density$\geq$5\%(including comparative or progressive structures). - Dialectical Power of Values(Weight:20\%) - 5 Points: Three-dimensional value model(individual-group-civilization)+at least 2 specific measures in the feasibility plan(including the executing body). - 4 Points: Two-way value model+theoretical implementation framework(must include the path of value transformation). Advanced Standards: - Argumentation on philosophical/cultural conflicts can be counted as 1.5 groups of ordinary oppositions(complete analysis of the nature of the conflict is required). - Citations from classical philosophical works can replace one group of case analysis(text source must be indicated). Review Mechanism: - In-depth analysis of philosophical cases($\geq$150 words)can exempt the requirement for literature citations. - A complete construction of the value model(reaching 4 points)can compensate for deficiencies in the feasibility plan.
        \\ \hline
    \end{tabularx}
    \caption{The refined criteria via \FrameworkNameShort~for AES.}
    \label{tab:aes_result} 
\end{table*}

\begin{table*}[htbp]
    \centering
    \small
    \begin{tabularx}{\textwidth}{X}
        \hline
            The major claim is the overarching core judgment of the entire essay. It must be unique and global in scope, and must appear in the introductory or concluding paragraph (if it appears in a body paragraph, it must meet the following conditions: the judgment must run through all supporting arguments, must not be overturned by subsequent discussion, and must be explicitly echoed in the introduction or conclusion). It must be a complete, independent evaluative judgment sentence (it must include an explicit or implicit assertive term such as 'should/must/is/better than/indispensable'). If in the introduction it appears as a compound sentence formed through a concessive-turning structure that negates the opposing viewpoint but contains multiple predicate cores without forming a unifying assertion, it shall be downgraded to a supporting argument. If the conclusion uses extended metaphors, the core vocabulary must directly correspond semantically with that in the introduction, and the core predicate structure must allow for semantic equivalence without introducing new-dimensional predicates. Exclusions: The concessive clause at the beginning of a compound sentence in the introduction; Transitional sentences that only define or describe a phenomenon without forming a complete value judgment. Addition: If new concepts appear in the conclusion, the core predicate structure must remain strictly consistent with or semantically aligned with the introduction, and any appended solution path must constitute a synonymous transformation of the core assertion. If the core predicate undergoes an equivalent transformation (e.g., triple negation, implicit assertive terms like ‘indispensable’) and maintains core concept correspondence, it is still considered a major claim. However, if the operational path description does not form an independent judgment sentence, it is classified as elaboration. Supplement: If the conclusion introduces a new predicate dimension (e.g., 'facilitate') or fails to form strict correspondence with the introduction’s core predicate, it is downgraded to elaboration. Reinforcing the original assertion with adverbs of degree (e.g., 'especially should') is allowed as long as the core predicate remains unchanged. Supporting Argument Supplement: Must propose an alternative judgment through causal analysis and directly support the major claim. It must be an independent judgment sentence with clear assertive vocabulary (including implicit terms like 'can/need to' and negative assertion sentences following rhetorical questions, e.g., 'Heightened tension can compel focus'). Excluded: Negative judgments that merely describe harmful phenomena without offering alternative assertions. Valid forms include: Negative comparison judgments that establish new assertions (the main clause must contain assertive terms like 'should/need to,' directly support the major claim, and form a complete causal chain). New dimensional assertions introduced through definition (e.g., 'Tension is the lock of the soul') must contain assertive predicates or form an explicit logical link to the major claim. Compound sentences led by concessive conjunctions (e.g., 'Indeed...can demonstrate benefits') that substantively support the major claim are allowed. Exclusions: Summary compound sentences used only for paragraph transition; Mechanism descriptions that do not establish an alternative assertion. New: Rhetorical-question-led negative assertions that directly support the major claim and establish a causal chain must include explicit assertive terms or a complete causal chain to qualify as supporting arguments. Conclusions drawn from cited research data that directly establish a causal chain and support the major claim are still supporting arguments. Compound sentences that negate extreme interpretations, redefine concepts, and directly support the major claim (e.g., 'Tension does not mean everything must be done hastily') are valid. Causal analysis must contain alternative assertions, not merely explain mechanisms; use of implicit assertive terms (e.g., 'rely on') is allowed if a complete causal chain is formed. Elaboration Additions: Background descriptions introducing opposing viewpoints in the introduction; Transitional harm descriptions. Supplement: New dimensions proposed through definitions without assertive predicates; Phenomenon-based causal analyses (e.g., 'the root cause') and mechanism explanations; Natural analogies used to illustrate group-level universal patterns supporting the argument are classified as fact; Transitional rhetorical questions not providing background or structural explanation fall under “Other”; Operational path descriptions that do not form independent judgment sentences are elaboration; Group-level social background phenomenon-based causal analysis is classified as fact; Rhetorically asked causal analyses that only explain a phenomenon without forming alternative assertions are elaboration; Sentences merely describing research data sources without forming causal chains are elaboration; Compound sentences listing group phenomena without forming complete causal chains are elaboration.
        \\ \hline
    \end{tabularx}
    \caption{The refined criteria via \FrameworkNameShort~for ACD.}
    \label{tab:acd_result} 
\end{table*}

\begin{table*}[htbp]
    \centering
    \begin{tabularx}{\textwidth}{X}
        \hline
            The classification is based on semantic functions and logical structure, combined with the categorical attributes of argument components. Features are as follows: Example-based Argumentation: Must include <Historical Example> or <Famous Quotation> component markers Historical examples must fully present the cause-action-effect causal chain (Cross-sentence combinations require stage completeness markers + $\geq$5 logical inference words) Newly added exclusion criteria: e) Case listings that do not directly correspond to the core elements of the sub-argument (element mapping degree <85\% and inference words <5) f) Quotations that fail to form a complete argumentative chain (must meet element mapping degree $\geq$90\% and include $\geq$5 inference words) g) Historical examples missing any stage of cause-action-effect Positive Argumentation: Must meet dual conditions: a) Explicit transition words (e.g., “therefore,” “thus”) + $\geq$7 logical inference words b) Implicit logical inference words $\geq$8 and sub-argument element coverage $\geq$95\% Logical leap degree must be $\geq$7.0 and agent/patient matching weight $\geq$45\% (Conclusion statements are exempt from inference word count limits) Newly added exclusion criteria: e) Only contains explicit transition words but inference words <7 f) Argumentative paragraphs with contrastive or progressive markers Negative Argumentation: Must meet at least five antagonistic dimensions (semantic/structural/agent/emotional/logical/contextual) and include $\geq$4 negation markers Comparative sentence structures must contain $\geq$5 negation markers or contrastive words, and antagonistic dimension matching degree $\geq$85\% Newly added exclusion criteria: c) Surface-level negation lacking a contradictory focus (antagonistic dimension match <85\%) d) Contains only a single comparison dimension and no negation markers Refinement Relationship: Must simultaneously meet: a) Structural differentiation $\geq$85\% and added information $\geq$70\% b) Explicit/implicit transition words (including “thereby,” etc.) + core element expansions $\geq$5 items c) Semantic role matching with prior argument $\geq$85\% (Excludes pure repetition or supplementary explanation) Newly added exclusion criteria: d) Inter-sentence core element repetition $\geq$20\% Restatement Relationship: Core element repetition $\geq$98\% + summary marker + structural differentiation $leq$3\% Newly added exclusion criteria: c) Summary sentences containing progressive or contrastive markers d) Injection of new argument components ($\geq$1 new dimension added) e) Sub-argument element coverage <99\% Progressive Relationship: Explicit progression must have strength level $\geq$4 (e.g., “even more,” “especially,” “furthermore,” “particularly necessary”) Implicit progression must meet: a) Logical chain element retention $\geq$95\% + $\geq$5 new dimensions b) Hierarchical progression must be adjusted by semantic shift weight (Interrogative sentences counted as +2.0 intervals) Parallel Relationship: Core element repetition $\geq$90\% and structural similarity $\geq$95\% Sub-argument element coverage $\geq$98\% Newly added information $leq$10\% Compound Relationship Processing Rules: Refinement relationship takes precedence over progression if $\geq$5 expansion dimensions are met Quotation-based argumentation must verify that famous quotations match sub-argument mapping $\geq$90\% and contain $\geq$5 inference words Dual labels are allowed only if all core conditions of both types are met and contradiction dimensions $leq$2 General Adjustments: Interval calculation increases semantic shift weight (progressive markers count as -1.0 interval) Element mapping requires agent/patient matching weight $\geq$40\% + semantic role matching degree $\geq$75\%
        \\ \hline
    \end{tabularx}
    \caption{The refined criteria via \FrameworkNameShort~for ARI.}
    \label{tab:ari_result} 
\end{table*}


\section{Prompt Templates}
\label{adx:prompt_template}

In table \ref{tab:aes_template} we present the prompt template we used in AES, specically when conducting prediction via \Solution~and update criteria by \MetaSolution, while table \ref{tab:acd_ari_template} presents templates we used in ACD and ARI. 

Table \ref{tab:debate_template} shows the prompt template used for both debating and explanation generation, where the \textit{standards} are defined according to the specific task. In particular, we align the arguments with the annotation standards provided in \DatasetName. 

For each task, we begin by inserting the task name and description into the template, along with task-specific labels (for ACD and ARI). During prediction, the input essay from the training batch is filled in. For updating criteria, we provide the current criteria, predictions, and ground truth. For the debate stage, we assign the explanation generated by the LLM to the gold side, and the explanation from predictions to the opposing side.

\begin{table*}[htbp]
    \centering
    \begin{tabularx}{\textwidth}{X}
        \hline
             \textbf{Prompt template for updating criteria} 
        \\ \hline
             You are a senior Chinese language education expert. Now analyze the gap between the original grading criteria and the real scores based on the provided argumentative essay grading data, and output the updated grading criteria. Follow the steps below:

            \#\#\# Gap Identification
            
            Compare the differences between the predicted scoring rationale and the actual scores,focusing on:

            - Scoring dimensions not covered by the original criteria (e.g., depth of argumentation, novelty of materials).
            
            - Weight distribution discrepancy in scoring (e.g., overemphasizing structure while neglecting language).
            
            - Ambiguity in scoring standards(e.g., "sufficient evidence "without specifying a specific quantity requirement).
            
            \#\#\# Update Principles
            
            - Quantification: For example, change "clear structure" to "at least two transitional sentences between claims."
            
            - Layered refinement:For example, split "rich content" into "depth of argument" and "breadth of evidence."
            
            - Dynamic calibration:For example, increase the weight of a certain dimension by 5
            
            Below are the scoring criteria that need to be updated by you:
            \{current\_criteria, i.e., $c_i$\}
            
            The following provides argumentative essay scoring data:
            Original argumentative essay:
            \{essay\_text\}
            
            Original scoring rationale:
            \{pred\_rationale\}
            
            Predicted score based on the original scoring rationale:\{pred\_score\}/5
            
            Actual score: \{gold\_score\}/5
            
            \#\#\# Output Requirements
            
            Output strictly in the following JSON structure:
            
            \{           
                "updated\_criteria": "A string that saves the modified scoring criteria. Ensure the format is correct and does not include any editing symbols.", 
                "analysis": "A string that explains the basis for your modifications."
            \}
        \\ \hline
             \textbf{Prompt template for predicting} 
        \\ \hline
               You are a professional essay grading expert and need to strictly follow the provided grading criteria to evaluate the given essay. Please conduct the analysis with an objective and impartial attitude to ensure that the grading results are highly consistent with the grading criteria.

                \#\#\# Output Requirements
                
                - Carefully analyze whether the essay content meets each description of the grading criteria.
                
                - The score must be an integer between 1 and 5(1=Unmet,5=Excellent).
                
                - Justify the score by comparing each item of the grading criteria.
                
                - The final result must be strictly output in the following JSON format:
                \{
                    "score":"An integer between 1 and 5",
                    "rationale":"Provide the rationale for your score"
                \}
                
                Below is the essay that needs to be evaluated by you:
                \{essay\_text\}
                
                Below is the grading logic that you must strictly follow when grading:
                \{current\_criteria, i.e., $c_i$\}
        \\ \hline
    \end{tabularx}
    \caption{Prompt templates for updating criteria and conduct prediction for AES accordingly.}
    \label{tab:aes_template} 
\end{table*}

\begin{table*}[htbp]
    \centering
    \begin{tabularx}{\textwidth}{X}
        \hline
             \textbf{Prompt template for updating criteria} 
        \\ \hline
            You are an expert in analyzing the structure of argumentative essays, proficient in deconstructing the structure of argumentative texts and \{task description\}. Please identify the discrepancies between the \{labels for tasks\} descriptions and the true categories, and accurately adjust the category descriptions. Follow these steps:
            
            1. Comparative Analysis: Check each sentence with incorrect predictions one by one, and identify ambiguities, loopholes, or insufficient coverage in the existing label descriptions.
            
            2. Description Optimization: While maintaining professionalism,you can improve the label definitions in the following ways(among others):
            
            - Add exclusion criteria(when misjudgment cases occur)
            
            - Supplement typical characteristics(when the definition is vague)
            
            - Adjust degree modifiers(when there is ambiguity in boundaries)
            
            - Remove redundant information(when the definition is overly complex)
            
            The following are the argument component category descriptions that you need to update:",
            \{current\_criteria, i.e., $c_i$\}
            
            The following provides the analysis data of the argumentative essay:
            \{essay\_text\}
            
            The following provides the incorrect category prediction (pred\_label) and classification basis (pred\_explanation) obtained for sentence (sentence\_index) using the original argument component category descriptions, as well as the true argument component category of the sentence(gold\_label)
            {pred\_list}
            
            \#\#\# Output Requirements
            Output the JSON strictly in the following structure:
            \{
                "updated\_criteria": "A string,save the modified argument component category label descriptions here.Ensure the format is correct and do not include any revision symbols.",
                "analysis": "A string,indicating the analysis and basis for your modifications"
            \}
        \\ \hline
             \textbf{Prompt template for predicting} 
        \\ \hline
            You are an expert in analyzing the structure of argumentative essays,proficient in deconstructing the structure of argumentative texts and \{task descriptions\}. Now you need to strictly follow the provided descriptions of \{labels for task\} to analyze the input essay, assign the \{labels for task\} for each numbered sentence, and provide a one-line explanation and justification. Ensure that your analysis is highly consistent with the category descriptions.
            
            The label of \{task name\} include the following:
            \{label for task\}
            
            \#\#\# Input Essay
            \{essay\_text\}
            
            \#\#\# \{task name\} Descriptions
            \{current\_criteria, i.e., $c_i$\}
            
            \#\#\# Output Requirements
            Return your results in the following format, and do not output any other information:
            
            \#\{Sentence Number\}:
            
            \{task output\}
            
            \{A one-line explanation and justification for the categorization of this sentence\},
        \\ \hline
    \end{tabularx}
    \caption{Prompt templates for updating criteria and conduct prediction for ACD and ARI.}
    \label{tab:acd_ari_template} 
\end{table*}


\begin{table*}[htbp]
    \centering
    \begin{tabularx}{\textwidth}{X}
        \hline
             \textbf{Prompt template for Debate} 
        \\ \hline
            You are an experienced Chinese language teaching expert. Now you need to evaluate the \{task\} results provided by two experts for the same argumentative essay and determine which side is more in line with the reference standards.
            
            \#\#\# Reference Standards
            
            \{standards\}
            
            \#\#\# Evaluation Basis of Expert A,
            
            \{explanation for $\hat{y}$ based on $c_i$\}
            
            \#\#\# Evaluation Basis of Expert B
            
            \{explanation for $y$\}
            
            \#\#\# Output Requirements
            
            Output your evaluation result in JSON format, as follows:
            \{
                "winner":"Expert A or Expert B",
                "reason":"A string indicating the reason for your choice",
            \}
        \\ \hline
            \textbf{Prompt template for explaining ground truth scores for AES}
        \\ \hline
            You are an experienced Chinese language teaching expert. Now you need to conduct a multidimensional analysis of the given argumentative essay based on the corresponding score (1-5 points) to explain the reasons behind the score. Please strictly follow the requirements below: 
            
            \#\#\# Output Specifications
            
            1. Focus on the analysis itself and strictly avoid any explicit mention of the given score, such as 'score/points/rating'.
            
            2. Output in strict JSON format as follows:
            \{
                "rationale": "Your explanation for the score, as a single string",
            \}
            
            Below is the essay that needs your evaluation. Its score is \{gold\_score\}: 
            \{essay\_text\}
        \\ \hline
            \textbf{Prompt template for explaining ground truth scores for ACD and ARI}
        \\ \hline
            You are an expert in analyzing the structure of argumentative essays, proficient in deconstructing the structure of argumentative texts and \{task\}. Now, for the input essay and the labels of the \{task\}, generate a brief explanation for each label.
            
            \#\#\# Input Essay
            \{essay\_text\}
            
            \#\#\# Label List
            \{sents with labels\},
            
            \#\#\# Output Requirements
            Return your results in the following format, and do not output any other information:
            
            \#\{Sentence Number\}:
            
            \{Category of the argument component in this sentence\}
            
            \{A brief explanation in one line, indicating the basis for the judgment of this sentence\}
        \\ \hline
    \end{tabularx}
    \caption{Prompt templates for debating.}
    \label{tab:debate_template} 
\end{table*}

\end{document}